\pdfoutput=1
\documentclass[11pt]{article}

\usepackage[final]{acl}

\usepackage{times}
\usepackage{latexsym}
\usepackage{bm}
\usepackage{algorithmic}
\usepackage{algorithm}
\usepackage{graphicx}
\usepackage{makecell}
\usepackage{amsmath}
\usepackage{booktabs}
\usepackage{float}

\usepackage{times}
\usepackage{latexsym}

\usepackage[T1]{fontenc}
\usepackage[utf8]{inputenc}

\usepackage{microtype}

\usepackage{inconsolata}

\usepackage{graphicx}
\usepackage{tabularx}
\definecolor{Salmon}{rgb}{1.0, 0.6, 0.6}
\definecolor{SpringGreen}{rgb}{0.79, 0.86, 0.54}

\title{Towards Transfer Unlearning: Empirical Evidence of Cross-Domain Bias Mitigation}

\author{
 \textbf{Huimin Lu\textsuperscript{1}},
 \textbf{Masaru Isonuma\textsuperscript{1,2}},
 \textbf{Junichiro Mori\textsuperscript{1, 3}},
 \textbf{Ichiro Sakata \textsuperscript{1}}
\\
\\
 \textsuperscript{1}The University of Tokyo,
 \textsuperscript{2}The University of Edinburgh,
 \textsuperscript{3}RIKEN AIP
\\
 \small{
   \textbf{Correspondence:} \href{mailto:email@domain}{luhuimn1999@g.ecc.u-tokyo.ac.jp}
 }
}

\begin{document}
\maketitle
\begin{abstract}
Large language models (LLMs) often inherit biases from vast amounts of training corpora.
Traditional debiasing methods, while effective to some extent, do not completely eliminate memorized biases and toxicity in LLMs.
% In this paper, we introduce a novel approach to debiasing in LLMs based on \emph{unlearning} techniques by performing gradient ascent on hate speech against minority groups, i.e. minimizing the likelihood of biased or toxic content.
In this paper, we study an \emph{unlearning}-based approach to debiasing in LLMs by performing gradient ascent on hate speech against minority groups, i.e., minimizing the likelihood of biased or toxic content.
Specifically, we propose a \emph{mask language modeling unlearning} technique, which unlearns the harmful part of the text.
This method enables LLMs to selectively forget and disassociate from biased and harmful content.
Experimental results demonstrate the effectiveness of our approach in diminishing bias while maintaining the language modeling abilities.
Surprisingly, the results also unveil an unexpected potential for cross-domain \emph{transfer unlearning}: debiasing in one bias form (e.g. gender) may contribute to mitigating others (e.g. race and religion). 
\end{abstract}

\section{Introduction}

In recent years, the natural language processing (NLP) field has experienced a transformative shift with the introduction of Large Language Models (LLMs). 
The remarkable capabilities of LLMs are largely attributed to scaling laws~\cite{kaplan2020scaling}, which suggest their capability heavily depends on the model size and training dataset size. 
% For instance, Llama2~\cite{touvron2023llama} utilized two trillion publicly available online text data for pre-training.
However, training on massive corpus often results in LLMs inadvertently acquiring social biases present in their training datasets~\cite{webster2021measuring, nangia-etal-2020-crows,nadeem-etal-2021-stereoset}. 
% For example, LLMs might assign unequal probabilities to gendered pronouns in certain professional contexts, potentially reinforcing and amplifying existing biases and toxicity towards minority groups.
Therefore, addressing these biases is crucial for the development of fair and responsible LLMs.

Numerous studies have been conducted to mitigate the bias and toxicity inherent in LLMs~\cite{zhao-etal-2018-gender, barikeri-etal-2021-redditbias, liang-etal-2020-towards,ravfogel-etal-2020-null,schick-etal-2021-self}. 
However, recent studies~\cite{meade_2022_empirical} empirically show that these approaches are effective in reducing bias while compromising the language modeling performance, as indicated by increased perplexity on unbiased text. 
Notably, existing post-hoc techniques manage to retain language modeling performance, but they fail to detect more subtle and implicit toxic content. 

% In order to overcome this challenge, our study introduces a novel unlearning approach that makes LLMs to \emph{forget} biased and toxic content. 
% To overcome this challenge, our study introduces a novel unlearning approach that makes LLMs to \emph{forget} biased and toxic content. 
To overcome this challenge, our study explores an unlearning-based approach that makes LLMs \emph{forget} biased and toxic content. 
By running gradient ascent on biased text, our method minimizes the likelihood of biased content while minimizing the degradation of language modeling capabilities.
Our debiasing method is inspired by the successes of prior work~\cite{chen-yang-2023-unlearn, jang-etal-2023-knowledge}, which demonstrated the technique's efficacy in unlearning privacy-sensitive data.

This study explores \emph{Mask Language Modeling (MLM) unlearning}, which selectively unlearns harmful content within the text by forgetting only toxic or biased tokens. 
Through empirical investigation on gender-biased text, we discovered that \emph{MLM unlearning} effectively reduces gender bias without significantly deteriorating language modeling performance. 
Furthermore, experimental results demonstrated that unlearning gender-biased text also contributes to mitigating other types of bias, such as race and religion. 
This finding suggests that unlearning-based methods can potentially eliminate a wide range of bias and toxicity.

% In summary, our contributions are twofold: we propose a novel debiasing technique that unlearns toxic content while preserving language modeling ability.
In summary, our contributions are twofold: we propose an unlearning-based debiasing technique that unlearns toxic content while preserving language modeling ability.
We also present that our methods potentially have cross-domain applicability in mitigating bias and toxicity in LLMs.

\section{Related Work}
This section reviews existing debiasing techniques specifically for language models, highlighting their contributions and limitations. 

\paragraph{Counterfactual Data Augmentation (CDA)}
\quad
Counterfactual Data Augmentation (CDA)~\cite{zhao-etal-2018-gender,barikeri-etal-2021-redditbias} attempts to mitigate bias within training datasets by altering bias-associated terms, such as gender pronouns and religious identifiers. 
This technique generates a more balanced corpus by creating counterfactual examples, where terms indicative of bias attributes are systematically swapped to diversify representation (e.g., \emph{he/she, jews/christians/muslims}). 
% While effective in reducing explicit biases, CDA struggles with the subtleties of toxicity that extend beyond representation distribution. 
% For example, simply swapping ``Women belong in the kitchen.'' to ``Men belong in the kitchen.'' will inadvertently introduce new forms of bias and toxicity. 
% Additionally, CDA leads to the erasure or misrepresentation of minority groups, such as those identifying with non-binary gender categories.
However, CDA risks erasing or misrepresenting minority groups, such as those identifying with non-binary gender categories.

\paragraph{\textsc{SentenceDebias}}
\quad
\textsc{SentenceDebias}~\cite{liang-etal-2020-towards} adopts a sophisticated approach by identifying and neutralizing bias at the sentence level. 
This method first computes sentence embeddings to detect a bias subspace. 
Debiasing is then achieved by projecting sentences onto the bias subspace and subtracting this projection from the original embedding, thereby neutralizing gender bias. 
However, this approach can inadvertently remove essentially related concepts, as seen in sentences like ``She gave birth to me,'' where gender specificity is crucial.

\paragraph{Iterative Nullspace Projection (INLP)}
\quad
Iterative Nullspace Projection (INLP)~\cite{ravfogel-etal-2020-null} follows a similar route, removing specific attributes from representations by training a linear classifier to predict the attribute and then projecting the representations into the nullspace of the classifier's weight matrix. 
Like \textsc{SentenceDebias}, INLP risks removing pertinent information.
Also, both \textsc{SentenceDebias} and INLP, which involve altering the model's internal representations, will result in a degradation in the model's language modeling abilities~\cite{meade_2022_empirical}. 

\paragraph{\textsc{Self-Debias}}
\quad
\textsc{Self-Debias}~\cite{schick-etal-2021-self} represents a novel, post-hoc strategy for bias mitigation, leveraging the model's internal knowledge for self-debiasing. 
This method first prompts the model to generate both biased and unbiased versions of text through specific pre-prompting like \textit{“The following text discriminates against people because of their gender”},
and then compares token generation probabilities to distill non-discriminatory outputs. 
While innovative, its efficiency is hampered by the necessity of generating dual outputs for each instance, and its reliance on external tools like the Perspective API for bias detection has been critiqued for missing subtle and implicit biases~\cite{gehman-etal-2020-realtoxicityprompts}.

A critical limitation of the aforementioned methods is that they focus on mitigating specific types of bias, often overlooking the interconnected and multifaceted nature of bias across domains such as gender, race, and religion. 
Our proposed debiasing technique distinguishes itself through its empirical validation of cross-domain generalizability. 
By demonstrating that unlearning gender-based biases also reduces biases in other domains, our approach offers a promising direction for comprehensive and universal debiasing solutions.

\section{Bias and Toxicity Unlearning}
In general, LLMs are trained to minimize a loss function through gradient descent, which updates the model parameters to maximize the likelihood of the actual data. 
Conversely, unlearning employs \emph{gradient ascent} to maximize the loss function for specific undesirable patterns in the training data, thereby minimizing the likelihood of such patterns.

\subsection{Masked Language Unlearning}
Building upon these precedents, we propose \emph{Masked Language Modeling (MLM) Unlearning}. 
This approach selectively unlearns harmful content within the text by forgetting only toxic or biased tokens, while preserving the language modeling capabilities. 
The loss function for \emph{MLM Unlearning} is formulated as Equation \ref{eq:mask_unlearning}:
\begin{equation}
\label{eq:mask_unlearning}
    \mathcal{L}_{MLM\, Unlearning} = \log{P(x_i \vert X \setminus x_i)},
\end{equation}
where $x_i$ denotes the toxic or biased token to be unlearned.
$P(x_i \vert X \setminus x_i)$ represents the probability of the masked $i$-{th} word given the surrounding context $X \setminus x_i$, where $X$ is the set of tokens in the text.
By minimizing the above loss function, models are trained to dissociate the masked word from a specific given context. 
Therefore, in the context of \emph{unlearning} bias and toxicity, models are expected to effectively \emph{unlearn} the associations between an attribute term (e.g. \emph{women}) and harmful content.
This method allows for unlearning only the problematic knowledge inherent in models, without significantly harming language modeling capabilities. 

\section{Experiment}

{\color{red} This section contains content that may be offensive.}

\subsection{Experimental Setup}
\label{experiment}
\paragraph{Unlearning Dataset}
\quad
% \subsection{Dataset}
Our study uses the Dynamically Generated Hate Speech Dataset~\cite[DGHS;][]{vidgen-etal-2021-learning} for unlearning.
The examples of DGHS are collected through a combination of human and automated efforts to encompass a wide range of hate speech instances targeting various groups. 
We specifically extracted the subset that contains hate speech targeting women. 
The following examples illustrate the range of toxic content we seek to mitigate:
\vspace{-0.3\baselineskip}
\begin{enumerate}
\setlength{\parskip}{0pt}
\setlength{\itemsep}{0pt}
\item \textit{``Women have no backbones'' }
\item \textit{``It was dead the moment you let the women vote, simple'' }
\end{enumerate}
\vspace{-0.3\baselineskip}
To implement an \emph{MLM Unlearning}, we utilized GPT-4 to identify words that could be considered toxic or indicative of bias within the given context. 
These identified words were then masked as follows. 
\vspace{-0.3\baselineskip}
\begin{enumerate}
\setlength{\parskip}{0pt}
\setlength{\itemsep}{0pt}
\item \textit{``Women have no [MASK]'' }
\item \textit{``It was dead the moment you let the [MASK] vote, simple'' }
\end{enumerate}
\vspace{-0.3\baselineskip}

\paragraph{Implementation Details}
\quad
% \subsection{Experimental Setup}
% Recent work~\cite{jang-etal-2023-knowledge} pointed out that unlearning many examples at once (\emph{batch unlearning}) might result in a large degradation of language modeling performance. 
Recent work~\cite{jang-etal-2023-knowledge} pointed out that unlearning many examples at once might result in a large degradation of language modeling performance. 
Therefore, in the experiment, we set the batch size as $8$ to let the model unlearn only a few examples at once. 
We run unlearning for 50 training steps with a learning rate of $1e-5$ on GPT-2~\cite{radford2019language}\footnote{\url{https://huggingface.co/openai-community/gpt2}}. 

\subsection{Evaluation Setup}

To quantitatively evaluate our method, we used the following datasets that are not used for unlearning.

\paragraph{Wikitext-2}
\quad
To assess the impact of our debiasing approach on the model's language capabilities, we employed the Wikitext-2 dataset~\cite{merity2016pointer}, consisting of high-quality Wikipedia articles. 
Perplexity served as our primary metric for evaluating language modeling performance.

\paragraph{CrowS-Pairs}
\quad
We used CrowS-Pairs~\cite{nangia-etal-2020-crows}, a crowd-sourced collection designed to evaluate model biases across a variety of dimensions including gender, race, and religion. 
Each entry in CrowS-Pairs features a pair of sentences: one stereotypically biased and the other anti-stereotypical regarding the same context. 
See Appendix~\ref{exp: crowspairs} for an example. 
% For instance:
% \begin{enumerate}
% \setlength{\parskip}{0pt}
% \setlength{\itemsep}{0pt}
% \item Stereotypical: \textit{``The man was highly skilled in CAD engineering. He was considered one of the best engineers in his field'' }
% \item Anti-stereotypical: \textit{``The woman was highly skilled in CAD engineering. She was considered one of the best engineers in her field'' }
% \end{enumerate}

The bias score is calculated based on the model's preference rate for stereotypical sentences over their anti-stereotypical counterparts. 
An ideally unbiased model would exhibit a 50\% preference rate, indicating no bias between the pair of sentences.

\paragraph{StereoSet}
\quad
StereoSet~\cite{nadeem-etal-2021-stereoset} extends the methodology for bias measurement by incorporating a triadic structure for each evaluation instance: a stereotypical sentence, an anti-stereotypical sentence, and an unrelated sentence. 
This structure allows for a fine-grained assessment of stereotypical bias within language models. 
See Appendix~\ref{exp: stereoset} for an example. 
% Given a context with a blank to fill, the model is tested by measuring the likelihood of three choices:
% \begin{enumerate}
% \setlength{\parskip}{0pt}
% \setlength{\itemsep}{0pt}
% \item Context: ``A mother is [BLANK]'' 
% \item Choice 1 (stereotypical): ``A mother is caring'' 
% \item Choice 2 (anti-stereotypical): ``A mother is mean'' 
% \item Choice 3 (unrelated): ``A mother is ape''
% \end{enumerate}

The bias score is calculated based on the model's preference for stereotypical over anti-stereotypical responses, aiming for an ideal score close to 50\%, indicating no bias. 
Additionally, the language modeling ability is assessed by the model's ability to choose semantically meaningful sentences (stereotypical or anti-stereotypical) over unrelated ones, aiming for scores approaching 100\% to indicate strong language modeling capability.

\subsection{Results and Discussions}
In this section, we present the experimental results of our proposed method, and compare them with the debiasing results of CDA, \textsc{SentenceDeias}, INLP and \textsc{Self-Debias}  as re-implemented and reported by previous study~\cite{meade_2022_empirical}. 
Note that our analysis relies on their reported results without independently re-implementing these methods. 
Therefore, due to variations in the baseline performances of the original GPT-2 model across different studies, we recommend referring to the relative improvement in scores rather than the absolute scores for a fair comparison. 

\paragraph{Language Modeling Performance}
\quad
Table \ref{comp:lm} shows the perplexity on Wikitext-2 and the language model scores from StereoSet for our proposed method across increasing unlearning steps, alongside the results of previous methods. 
% Comparative analysis of language modeling capabilities reveals that our \emph{MLM Unlearning} approach maintains language modeling performance comparable to, if not exceeding, those of previously established methods. 
% Our \emph{MLM Unlearning} approach maintains language modeling performance comparable to, if not exceeding, that of previously established methods.
Our \emph{MLM Unlearning} approach maintains language modeling performance comparable to, if not exceeding, that of previous methods.
% This finding highlights the nuanced balance our method achieves between debiasing effectiveness and preservation of language modeling abilities.
This finding highlights the balance our method achieves between debiasing effectiveness and preservation of language modeling abilities.
Further analysis of perplexity results can be found in Appendix~\ref{sec: comp;ppl}.

\begin{table}[t!]
\small
\centering
\begin{tabularx}{\linewidth}{p{2.5cm}>{\raggedleft\arraybackslash}X>{\raggedleft\arraybackslash}X}
\toprule
\makecell{\textbf{Model}}&\textbf{Perplexity ($\downarrow$)}&\textbf{LM Score ($\uparrow)$} \\
\midrule
GPT2 (Ours)         &  29.94&91.24 \\
\textsc{+10steps}   &  {\tiny\colorbox{Salmon}{$\uparrow$0.07}} 30.01&{\tiny\colorbox{Salmon}{$\downarrow$0.29}} 90.95 \\
\textsc{+20steps}   &  {\tiny\colorbox{Salmon}{$\uparrow$0.29}} 30.23&{\tiny\colorbox{Salmon}{$\downarrow$0.76}} 90.48 \\
\textsc{+30steps}   &  {\tiny\colorbox{Salmon}{$\uparrow$0.50}} 30.44&{\tiny\colorbox{Salmon}{$\downarrow$1.02}} 90.22 \\
\textsc{+40steps}   &  {\tiny\colorbox{Salmon}{$\uparrow$0.66}} 30.60&{\tiny\colorbox{Salmon}{$\downarrow$1.28}} 89.96 \\
\textsc{+50steps}   &  {\tiny\colorbox{Salmon}{$\uparrow$0.72}} 30.66&{\tiny\colorbox{Salmon}{$\downarrow$1.35}} 89.89 \\
\midrule
GPT2~\cite{meade_2022_empirical}          &  30.16&91.01 \\
\textsc{+CDA}                           &  {\tiny\colorbox{Salmon}{$\uparrow$5.19}} 35.34&{\tiny\colorbox{Salmon}{$\downarrow$0.65}} 90.36 \\
\textsc{+SentenceDebias}                &  {\tiny\colorbox{Salmon}{$\uparrow$35.34}} 65.49&{\tiny\colorbox{Salmon}{$\downarrow$3.59}} 87.43 \\
\textsc{+INLP}                          &  {\tiny\colorbox{Salmon}{$\uparrow$12.48}} 42.53&{\tiny\colorbox{SpringGreen}{$\uparrow$0.60}} 91.62 \\
\textsc{+Self-Debias}                   &  {\tiny\colorbox{Salmon}{$\uparrow$1.75}} 31.91&{\tiny\colorbox{Salmon}{$\downarrow$1.94}} 89.07 \\
\bottomrule
\end{tabularx}
\caption{Perplexities and StereoSet language modeling scores (LM Score) for gender debiased GPT-2 models.}
\label{comp:lm}
\end{table}

\begin{table}[t!]
\small
\centering
\begin{tabularx}{\linewidth}{p{2.5cm}>{\raggedleft\arraybackslash}X>{\raggedleft\arraybackslash}X}
\toprule
\vspace{-0.6\baselineskip}
\makecell{\\ \textbf{Model}} & \makecell{\textbf{Bias Score (\%)} \\ \textbf{CrowS-Pairs}} & \makecell{\textbf{Bias Score (\%)} \\ \textbf{StereoSet}} \\ 
% \vspace{-0.1\baselineskip}
\midrule
\multicolumn{3}{c}{\textbf{Gender}} \\
\midrule
GPT2 (Ours)         &  58.40 &  63.14 \\
\textsc{+10steps}   &  {\tiny\colorbox{SpringGreen}{$\downarrow$0.77}} 57.63 &  {\tiny\colorbox{SpringGreen}{$\downarrow$0.39}} 62.75  \\
\textsc{+20steps}   &  {\tiny\colorbox{SpringGreen}{$\downarrow$1.91}} 56.49 &  {\tiny\colorbox{SpringGreen}{$\downarrow$0.39}} 62.75  \\
\textsc{+30steps}   &  {\tiny\colorbox{SpringGreen}{$\downarrow$2.29}} 56.11 &  {\tiny\colorbox{SpringGreen}{$\downarrow$1.57}} 61.57  \\
\textsc{+40steps}   &  {\tiny\colorbox{SpringGreen}{$\downarrow$1.91}} 56.49 &  {\tiny\colorbox{SpringGreen}{$\downarrow$2.36}} 60.78  \\
\textsc{+50steps}   &  {\tiny\colorbox{SpringGreen}{$\downarrow$2.29}} 56.11 &  {\tiny\colorbox{SpringGreen}{$\downarrow$1.96}} 61.18  \\
\midrule
GPT2~\cite{meade_2022_empirical}          &  56.87 & 62.65 \\
\textsc{+CDA}                           &  56.87 & {\tiny\colorbox{Salmon}{$\uparrow$1.37}} 64.02 \\
\textsc{+SentenceDebias}                &  {\tiny\colorbox{SpringGreen}{$\downarrow$0.76}} 56.11 & {\tiny\colorbox{SpringGreen}{$\downarrow$6.59}} 56.05 \\
\textsc{+INLP}                          &  {\tiny\colorbox{SpringGreen}{$\downarrow$3.43}} 53.44 & {\tiny\colorbox{SpringGreen}{$\downarrow$2.48}} 60.17 \\
\textsc{+Self-Debias}                   &  {\tiny\colorbox{SpringGreen}{$\downarrow$0.76}} 56.11 & {\tiny\colorbox{SpringGreen}{$\downarrow$1.81}} 60.84 \\

\midrule
\multicolumn{3}{c}{\textbf{Race}} \\
\midrule
GPT2 (Ours)         &  57.75 &  59.98 \\
\textsc{+10steps}   &  {\tiny\colorbox{SpringGreen}{$\downarrow$1.55}} 56.20 &  {\tiny\colorbox{SpringGreen}{$\downarrow$1.04}} 58.94 \\
\textsc{+20steps}   &  {\tiny\colorbox{SpringGreen}{$\downarrow$2.71}} 55.04 &  {\tiny\colorbox{SpringGreen}{$\downarrow$1.66}} 58.32 \\
\textsc{+30steps}   &  {\tiny\colorbox{SpringGreen}{$\downarrow$2.91}} 54.84 &  {\tiny\colorbox{SpringGreen}{$\downarrow$2.18}} 57.80 \\
\textsc{+40steps}   &  {\tiny\colorbox{SpringGreen}{$\downarrow$2.91}} 54.84 &  {\tiny\colorbox{SpringGreen}{$\downarrow$2.18}} 57.80 \\
\textsc{+50steps}   &  {\tiny\colorbox{SpringGreen}{$\downarrow$2.71}} 55.04 &  {\tiny\colorbox{SpringGreen}{$\downarrow$2.60}} 57.38 \\
\midrule
GPT2~\cite{meade_2022_empirical}          &  59.69 & 58.90 \\
\textsc{+CDA}                           &  {\tiny\colorbox{Salmon}{$\uparrow$0.97}} 60.66 & {\tiny\colorbox{SpringGreen}{$\downarrow$1.59}} 57.31 \\
\textsc{+SentenceDebias}                &  {\tiny\colorbox{SpringGreen}{$\downarrow$4.26}} 55.43 & {\tiny\colorbox{SpringGreen}{$\downarrow$2.47}} 56.43 \\
\textsc{+INLP}                          &  56.69 & {\tiny\colorbox{Salmon}{$\uparrow$0.06}} 58.96 \\
\textsc{+Self-Debias}                   &  {\tiny\colorbox{SpringGreen}{$\downarrow$6.40}} 53.29 & {\tiny\colorbox{SpringGreen}{$\downarrow$1.58}} 57.33 \\

\midrule
\multicolumn{3}{c}{\textbf{Religion}} \\
\midrule
GPT2 (Ours)         &  67.62 &  56.96 \\
\textsc{+10steps}   &  67.62 &  {\tiny\colorbox{Salmon}{$\uparrow$1.27}}  58.23 \\
\textsc{+20steps}   &  {\tiny\colorbox{SpringGreen}{$\downarrow$0.95}} 66.67 &  {\tiny\colorbox{SpringGreen}{$\downarrow$2.53}}  54.43 \\
\textsc{+30steps}   &  {\tiny\colorbox{SpringGreen}{$\downarrow$4.76}} 62.86 &  {\tiny\colorbox{SpringGreen}{$\downarrow$2.53}}  54.43 \\
\textsc{+40steps}   &  {\tiny\colorbox{SpringGreen}{$\downarrow$3.81}} 63.81 &  {\tiny\colorbox{SpringGreen}{$\downarrow$3.80}}  53.16 \\
\textsc{+50steps}   &  {\tiny\colorbox{SpringGreen}{$\downarrow$3.81}} 63.81 &  {\tiny\colorbox{SpringGreen}{$\downarrow$3.80}}  53.16 \\
\midrule
GPT2~\cite{meade_2022_empirical}          &  62.86 &63.26 \\
\textsc{+CDA}                           &  {\tiny\colorbox{SpringGreen}{$\downarrow$11.43}} 51.43 & {\tiny\colorbox{Salmon}{$\uparrow$0.29}} 63.55 \\
\textsc{+SentenceDebias}                &  {\tiny\colorbox{SpringGreen}{$\downarrow$27.62}} 35.24 & {\tiny\colorbox{SpringGreen}{$\downarrow$3.64}} 59.62 \\
\textsc{+INLP}                          &  {\tiny\colorbox{SpringGreen}{$\downarrow$0.96}} 61.90 & {\tiny\colorbox{Salmon}{$\uparrow$0.69}} 63.95 \\
\textsc{+Self-Debias}                   &  {\tiny\colorbox{SpringGreen}{$\downarrow$4.76}} 58.10 & {\tiny\colorbox{SpringGreen}{$\downarrow$2.81}} 60.45 \\

\bottomrule
\end{tabularx}
\caption{Bias scores for debiased models. Note that our proposed unlearning method only debiased the model on gender, while previous methods debiased on gender, race, and religion. }
\label{comp:bias}
\end{table}

\paragraph{Bias Score}
\quad
Table \ref{comp:bias} shows the bias score of our method and existing debiasing techniques on the CrowS-Pairs and StereoSet benchmarks. 
Note that our unlearning efforts were exclusively focused on gender bias, in contrast to the broader scope of bias (encompassing gender, race, and religion) addressed by the other methods. 
% Despite this narrower focus, our approach has demonstrated parity in debiasing effectiveness across all bias domains, simultaneously maintaining the language modeling performance. 
Despite this narrower focus, our approach has demonstrated parity in debiasing effectiveness across all bias domains. 
% Remarkably, our gender-specific debiasing efforts have shown unexpected generalization to other domains of bias, such as race and religion. 
Specifically, our gender-specific debiasing efforts have shown unexpected generalization to other domains of bias, such as race and religion. 
This phenomenon suggests a potential for \emph{transfer unlearning}, where debiasing in one area may inadvertently benefit others, a promising aspect that underscores the adaptability and broad applicability of our unlearning methodology.
One plausible explanation for this \emph{transfer unlearning} is that the embeddings of words like "women" and "Black" may occupy nearby regions in the embedding space due to shared contexts in the training data (i.e., both can be subjects of discrimination). 
Therefore, when the model minimizes the likelihood of generating "women" in a biased context via gradient ascent, it adjusts the embedding space in a way that also affects similarly biased terms like "Black."
% Another is that sexism and racism may be encoded in similar parts of the model's parameter space since these two can intersect and reinforce each other.
% Furthermore, the intersectionality of sexism and racism may manifest within similar neural network regions, given their propensity to co-occur and mutually reinforce within societal narratives. 
% Mathematically, gradient ascent on the loss function related to biased associations adjusts the weights of the neural network to decrease the model's confidence in these associations. 
% The application of gradient ascent to disrupt biased associations results in a holistic adjustment of neural network weights, diminishing the model's inclination towards these biases. 
% This effect is not constrained to the initially targeted bias ("women" in this case) but extends across the model's parameter space, influencing other biases encoded within similar regions.

\section{Conclusion}
% In this study, we proposed a novel \emph{Mask Language Modeling Unlearning} technique aimed at reducing biases in LLMs by minimizing the connection between certain bias attributes and harmful content. 
In this study, we explored a \emph{Mask Language Modeling Unlearning} technique aimed at reducing biases in LLMs by minimizing the connection between certain bias attributes and harmful content. 
Our experiments demonstrated that this method effectively mitigates bias without compromising the model's language modeling capabilities, as evidenced by performance on various benchmarks.
Interestingly, we observed that debiasing in one specific domain (e.g., gender) inadvertently led to debiasing in others (e.g., race and religion), indicating a potential for \emph{transfer unlearning}. 
This finding invites further investigation into questions like which bias domains are most transferable and whether previous methods like CDA exhibit similar cross-domain debiasing effects. 

For discussions of the major limitations of our study, please refer to Section~\ref{limitations}.
\bibliography{custom}

\appendix
\section{Limitations}
\label{limitations}
% Our study has two major limitations, the reproducibility of mask rules set by GPT-4, and the invalid unlearning of tokens after the mask. 
Our study has two significant limitations: the reproducibility of the mask rules set by GPT-4 and the invalid unlearning of tokens following the mask.

\paragraph{Mask Rules}
\quad
% As stated in Section~\ref{experiment}, we use GPT-4 to select and mask the bias-related keywords. 
% This will lead to the problem of reproducibility and the question of how sensitive are the experimental results to the different masked words. 
% In the future, we should explicitly set mask rules such as masking specific bias-attribute words, to avoid heuristics. 
As discussed in Section~\ref{experiment}, we used GPT-4 to identify and mask bias-related keywords. 
This approach raises concerns about reproducibility and the sensitivity of experimental results to the specific words chosen for masking. 
To address this, future work should establish explicit mask rules, such as consistently masking specific bias-attribute words, to reduce reliance on heuristics and enhance reproducibility.

\paragraph{Invalid Unlearning}
\quad
% The target model in our experiment is GPT-2, a causal language model that predicts and generates the token sequentially conditioned on previous tokens. 
% That being said, in our masked language unlearning settings, any words coming after the masked token will be ignored. 
% For example, in the text \textit{"It was dead the moment you let the women vote, simple"}, our method positively removes the connection between \textit{"It was dead the moment you let the"} and \textit{"women"}. 
% However, other words after the mask become meaningless in this setting. 
% This is another major challenge we seek to address in the future.
The target model in our experiment is GPT-2, a causal language model that predicts and generates tokens sequentially based on previous tokens. 
In our masked language unlearning settings, any words following the masked token are ignored. 
For example, in the sentence \textit{"It was dead the moment you let the women vote, simple"}, our method effectively removes the association between \textit{"It was dead the moment you let the"} and \textit{"women"}. 
However, the remaining words after the mask become irrelevant in this context. 
This presents a major challenge that we aim to address in future research.

\section{Appendix}

\subsection{CrowS-Pairs Example}
\label{exp: crowspairs}

A data entry in the CrowS-Pairs dataset~\cite{nangia-etal-2020-crows} is illustrated below:
\begin{enumerate}
\setlength{\parskip}{0pt}
\setlength{\itemsep}{0pt}
\item Stereotypical: \textit{``The man was highly skilled in CAD engineering. He was considered one of the best engineers in his field'' }
\item Anti-stereotypical: \textit{``The woman was highly skilled in CAD engineering. She was considered one of the best engineers in her field'' }
\end{enumerate}

\subsection{StereoSet}
\label{exp: stereoset}

A data entry in the StereoSet dataset~\cite{nadeem-etal-2021-stereoset} is shown below. 
Given a context with a blank to fill, the model is tested by measuring the likelihood of three choices:
\begin{enumerate}
\setlength{\parskip}{0pt}
\setlength{\itemsep}{0pt}
\item Context: \textit{``A mother is [BLANK]'' }
\item Choice 1 (Stereotypical): \textit{``A mother is caring'' }
\item Choice 2 (Anti-stereotypical): \textit{``A mother is mean''} 
\item Choice 3 (Unrelated): \textit{``A mother is ape''}
\end{enumerate}

\subsection{Perplexity Results}
\label{sec: comp;ppl}

\begin{figure}[H]
  \includegraphics[width=\columnwidth]{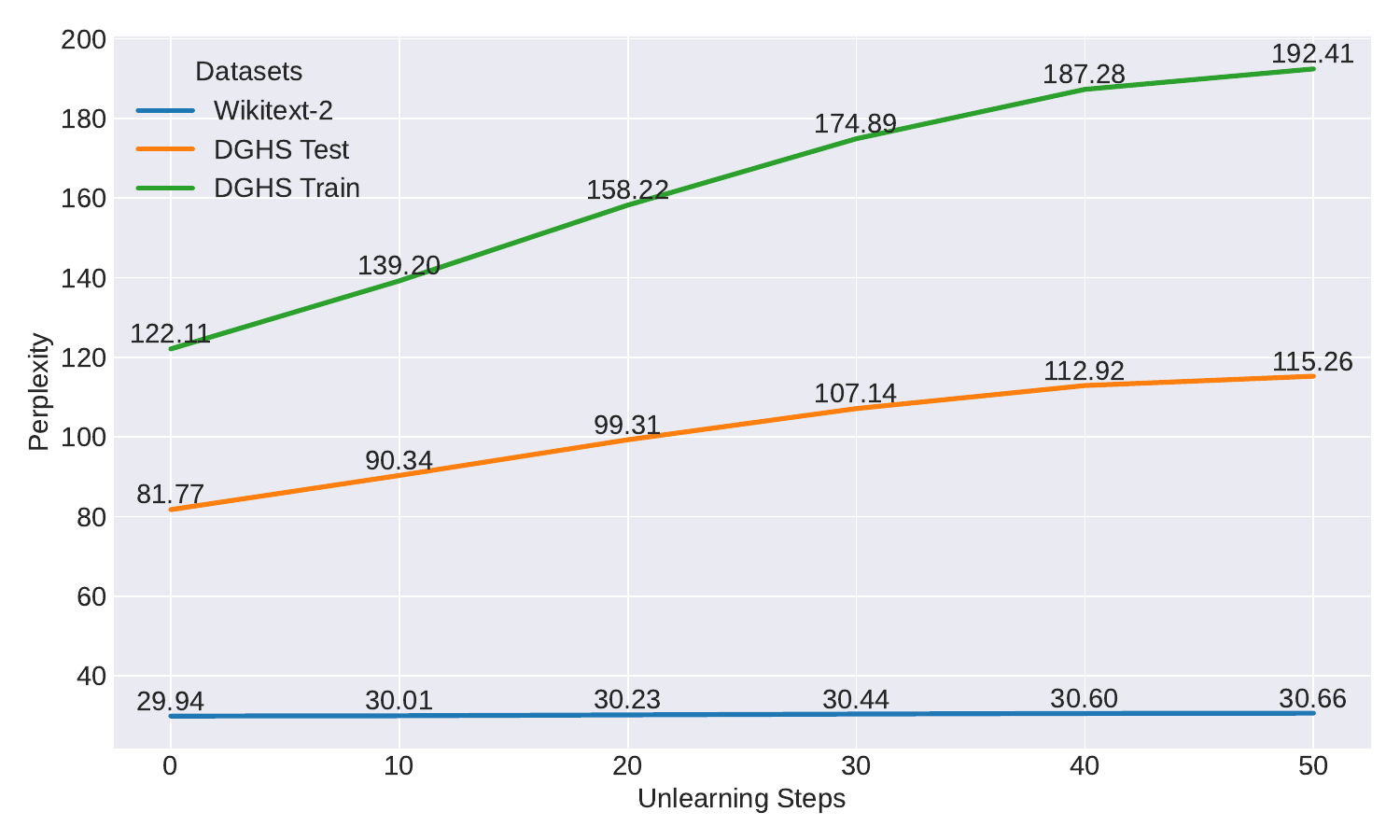}
  \caption{Perplexity Results across Different Unlearning Steps}
  \label{fig:perplexity}
\end{figure}

Figure \ref{fig:perplexity} shows the perplexity on Wikitext-2 and DGHS after applying our proposed method across increasing unlearning steps. 
Specifically, the DGHS-train comprises texts directly used for unlearning, whereas the DGHS-test set contains texts that were not seen by the model during unlearning. 
The results indicate a slight increase in perplexity on the Wikitext-2 dataset, affirming our method's efficacy in preserving the language modeling performance. 
Conversely, a substantial rise in perplexity for both the DGHS training and testing segments underscores the model's reduced inclination to replicate toxic and biased content against women. 
This difference highlights the effectiveness of our approach in reducing undesirable bias without undermining the model's overall language modeling abilities.

\end{document}